\documentclass[11pt, a4paper]{article}
\usepackage{coling2016}
\usepackage{latexsym}

\usepackage{times}
\usepackage{graphicx} 
\usepackage{times}    
\usepackage{url}      
\usepackage{algorithm}
\usepackage{algorithmic}
\usepackage{amsmath}
\usepackage{amssymb}
\usepackage{color}
\usepackage{caption}
\usepackage{enumerate}
\usepackage{subcaption}
\usepackage[table]{xcolor}
\usepackage{multirow}
\usepackage{CJKutf8}
\usepackage{multicol}
\usepackage{todonotes}

\usepackage[utf8]{inputenc}
\usepackage{tikz}
\usetikzlibrary{shapes,arrows}
\usepackage{soul}  

\definecolor{REd}{rgb}{0.5,0.2,0.2}
\definecolor{GREEn}{rgb}{0.2,0.4,0.2}
\definecolor{BLUe}{rgb}{0.2,0.2,0.6}

\title{Grammatical Templates: Improving Text Difficulty Evaluation for Language Learners}
\author{Shuhan Wang \\
  Department of Computer Science\\
  Cornell University \\
  {\tt forsona@cs.cornell.edu} \\\And
  Erik Andersen \\
  Department of Computer Science\\
  Cornell University \\
  {\tt eland@cs.cornell.edu} \\}

\date{}

\begin{document}
\maketitle

\begin{abstract}
Language students are most engaged while reading texts at an appropriate difficulty level. However, existing methods of evaluating text difficulty focus mainly on vocabulary and do not prioritize grammatical features, hence they do not work well for language learners with limited knowledge of grammar. In this paper, we introduce \emph{grammatical templates}, the expert-identified units of grammar that students learn from class, as an important feature of text difficulty evaluation. Experimental classification results show that grammatical template features significantly improve text difficulty prediction accuracy over baseline readability features by 7.4\%. Moreover, we build a simple and human-understandable text difficulty evaluation approach with 87.7\% accuracy, using only 5 grammatical template features. 

\end{abstract}

\vfill

\blfootnote{
    %
    %
    %
    %
    %
    %
     \hspace{-0.65cm}  
     This work is licenced under a Creative Commons 
     Attribution 4.0 International License.
     License details:
     \url{http://creativecommons.org/licenses/by/4.0/}
}

\section{Introduction}

Evaluating \emph{text difficulty}, or \emph{text readability}, is an important topic in natural language processing and applied linguistics~\cite{zamanian2012readability,pitler2008revisiting,fulcher1997text}. A key challenge of text difficulty evaluation is that linguistic difficulty arises from both vocabulary and grammar~\cite{richards2013longman}. 
However, most existing tools either do not sufficiently take the impact of grammatical difficulty into account~\cite{smith2014beyond,sheehan2014textevaluator}, or use traditional syntactic features, which differ from what language students actually learn, to estimate grammatical complexity~\cite{schwarm2005reading,heilman2008analysis,franccois2012ai}. In fact, language courses introduce grammar constructs together with vocabulary, and grammar constructs vary in frequency and difficulty just like vocabulary~\cite{blyth1997constructivist,manzanares2008can,waara2004construal}. Ideally, we would like to have better ways of estimating the grammatical complexity of a sentence.

To make progress in this direction, we introduce \emph{grammatical templates} as an important feature in text difficulty evaluation. These templates are what language teachers and linguists have identified as the most important units of grammatical understanding at different levels, and what students actually learn in language lessons. We also demonstrate that grammatical templates can be automatically extracted from the dependency-based parse tree of a sentence.

To evaluate, we compare the difficulty prediction accuracy of grammatical templates with existing readability features in Japanese language placement tests and textbooks. Our results show that grammatical template features slightly outperform existing readability features. Moreover, adding grammatical template features into existing readability features significantly improves the accuracy by 7.4\%.
We also propose a multilevel linear classification algorithm using only 5 grammatical features. We demonstrate that this simple and human-understandable algorithm effectively predicts the difficulty level of Japanese texts with 87.7\% accuracy.

\section{Related Work}

Text difficulty evaluation has been widely studied over the past few decades~\cite{nelson2012measures,sinha-EtAl:2012:POSTERS,hancke-vajjala-meurers:2012:PAPERS,jameel-qian-lam:2012:PAPERS,gonzalezdios-EtAl:2014:Coling,sinha-dasgupta-basu:2014:Coling}. Researchers have developed over 200 metrics of text difficulty~\cite{collins2004language}. For example, \emph{Lexile} measures text complexity and readability with word frequency and sentence length~\cite{smith2014beyond}. \emph{ATOS}\footnote{http://www.renaissance.com/Products/Accelerated-Reader/ATOS/ATOS-Analyzer-for-Text} includes two formulas for texts and books, both of which take into account three variables to predict text difficulty: word length, word grade level and sentence length. \emph{TextEvaluator} is a comprehensive text analysis system designed to help teachers and test developers evaluate the complexity characteristics of reading materials~\cite{sheehan2014textevaluator}. It incorporates more vocabulary features, such as meaning and word type, as well as some sentence and paragraph-level features.

Nevertheless, most of these methods provide limited consideration of grammatical difficulty, which is a major challenge for foreign language learners~\cite{callan2007combining}. In fact, text readability not only depends on sentence lengths or word counts, but on `the grammatical complexity of the language used' as well~\cite{richards2013longman}. Based on this fact, recent readability evaluation systems improved performance by incorporating syntactic features like parse tree depth~\cite{schwarm2005reading} and subtree patterns~\cite{heilman2008analysis} to measure grammatical complexity. Moreover, researchers have developed an unified framework of text readability evaluation, which combines lexical, syntactic and discourse features, and predicts readability with outstanding accuracy~\cite{pitler2008revisiting}. The relationship between text readability and reading devices was also studied in the past two years~\cite{kim2014device}. However, most of these approaches are intended for native speakers and use texts from daily news, economic journals or scientific publications, which are too hard to read for beginning and intermediate language learners. Ideally, we would have specific features and approaches for text difficulty evaluation for language learners.

Recently, language educational researchers conducted a bunch of studies on text readability evaluation for language learners in different languages, such as English, German, Portuguese and French~\cite{blyth1997constructivist,waara2004construal,manzanares2008can,vajjala2012improving,franccois2012ai,xia2016text}. However, they use traditional syntactic features such as sentence length, part of speech ratios, number of clauses and average parse tree height, which differ from the grammatical knowledge that students actually learn in language lessons. For example, Curto et al. measured text difficulty using traditional vocabulary and syntactic features, to predict text difficulty levels for Portuguese language learners~\cite{curto2015assisting}. Unfortunately, 75\% accuracy in 5-level classification with 52 features is not satisfactory. Instead, we extract grammatical features from \textit{grammatical templates}, the knowledge units that language students actually learn in classes and that expert language instructors have identified and highlighted in textbooks. We also propose a novel technique that has a simpler and human-interpretable structure, uses only 5 grammatical template features, and predicts text difficulty with 87.7\% accuracy in 5-level classification.

\begin{CJK}{UTF8}{min}

\section{Grammatical Template Analysis}

A key challenge in modeling text difficulty is to specify all prerequisite knowledge required for understanding a certain sentence. Traditional methods measure text difficulty mostly by evaluating the complexity of vocabulary (word count, word frequency, word type, etc.). This is effective for native speakers, who typically understand the grammar of their language but vary in mastery of vocabulary. However, these vocabulary-based methods underperform for language learners who have limited knowledge of grammar~\cite{callan2007combining,curto2015assisting}.

To resolve this, we focus our research on grammatical difficulty. We introduce the idea of \emph{grammatical templates}, units of grammar that expert language instructors and linguists have identified as the most important grammatical knowledge, and are typically emphasized as key points in every textbook lesson~\cite{Genki,StandardJapanese}. Since these grammatical templates are taught explicitly in language lessons and learned directly by language students, we believe they reflect the conceptual units of grammar more closely than parse trees. 

Grammatical templates play an important role in language understanding because:

\begin{itemize}
\item Many grammatical templates suggest sentence structure. For example, ``hardly ... when ...'' in English, ``nicht nur ..., sondern auch ...'' (not only ... but also ...) in German, and ``必ずしも ... とはいえない'' (it is not necessarily the case that ...) in Japanese;
\item For languages like Chinese and Japanese, lacking knowledge of some grammatical templates will cause difficulties in segmentation. For example, consider the Japanese template ``...つ...つ'' (two opposite behaviors occuring alternately) in the phrase ``行きつ戻りつ'' (to walk back and forth), and the Chinese template ``越...越好'' (the more ... the better) in ``越早越好''(the earlier the better);
\item Some grammatical templates may refer to special meanings that cannot be understood as the combination of individual words. For example, ``in terms of'', ``such that'' in English, ``mit etwas zu tun haben'' (have something to do with ...) in German, and ``... ことはない'' (no need to ...) in Japanese.
\end{itemize}

We show some simple examples of grammatical templates for Japanese in Table \ref{tab:templates}\footnote{A long list of Japanese grammatical templates with English translations can be accessed at the JGram website: http://www.jgram.org/pages/viewList.php. There is also a nice and comprehensive book of Japanese grammatical templates, written by Japanese linguists, with English, Korean and Chinese translations:~\cite{tomomatsu2010essential}.}. Line 2 shows the pronunciation of the templates, line 3 shows the translations, and the uppercase letters in line 4 are provided for notation. We also provide examples of how the grammar of a sentence can be described as combinations of these grammatical templates in Table \ref{tab:process}.

\begin{table*}

\centering
\begin{tabular}{|c|c|c|c|c|c|c|}
\hline
Template&-- は&-- の&-- を&-- では ない&--(名詞) に&--(動詞連用形) に \\
\hline
Pronunciation &-- \textit{wa}&-- \textit{no}&-- \textit{o}&-- \textit{dewa\ \ nai}&--(noun) \textit{ni}&--(verb, i-form) \textit{ni}\\
\hline
Translation & (topic) & (genitive) & (object) & is not & to (location)& for (purpose) \\
\hline
Notation&A&B&C&D&E&F\\
\hline
\end{tabular}
\caption{Grammatical Templates in Japanese, with hyphens denoting words to be filled in. Note that some grammatical templates may impose requirements of some properties (e.g. part of speech or form) on the missing words.
}
\label{tab:templates}


\vspace{0.15in}
\centering
\begin{tabular}{|llllllll|}
\hline
Sentence   &彼&\textbf{は}&すぐ&東京&\textbf{に}&到着する&\\
Pronunciation&kare&\textbf{wa}&sugu&\textit{toukyou}&\textbf{ni}&touchakusuru&\\
Translation&he  &\textbf{(topic)} &soon&\textit{Tokyo}&\textbf{to (location)}&arrive&\\
Templates  &&A&&&E&&\\
           &\multicolumn{5}{l}{`` he will soon arrive in Tokyo ''}&&\\
\hline
Sentence   &僕&\textbf{の}&彼女&\textbf{を}&見&\textbf{に}&行く\\
Pronunciation&boku&\textbf{no}&kanojo&\textbf{o}&\textit{mi}&\textbf{ni}&iku \\
Translation&I  &\textbf{(genitive)} &girlfriend&\textbf{(object)}&\textit{see}&\textbf{for (purpose)}&go\\
Templates  &&B&&C&&F&\\
           &\multicolumn{5}{l}{`` I go to see my girlfriend ''}&&\\
\hline
Sentence   &これ&\textbf{は}&君&\textbf{の}&本&\textbf{では}&\textbf{ない}\\
Pronunciation &kore&\textbf{wa}&kimi&\textbf{no}&hon&\textbf{dewa}&\textbf{nai}\\
Translation&this  &\textbf{(topic)} &you&\textbf{(genitive)}&book&\textbf{is}&\textbf{not}\\
Templates  &&A&&B&&D&\\
           &\multicolumn{5}{l}{`` this is not your book ''}&&\\
\hline
\end{tabular}

\caption{Identified grammatical templates of Japanese sentences. In sentences, pronunciations and translations, grammatical templates are in bold. The word \textit{toukyou} in the first sentence is a noun (Tokyo,東京), as characterized by template~E. The word \textit{mi} (to see, 見) in the second sentence is the i-form (動詞連用形) of a verb, as required by template~F.}
\label{tab:process}

\vspace{0.15in}

\begin{tabular}{|c|r|r|r|r|r|}
\hline
& N1 Texts & N2 Texts& N3 Texts& N4 Texts & N5 Texts \\
\hline
N1 Templates & 0.902\% & 0.602\% & 0.077\% & 0.074\% & 0.056\% \\
\hline
N2 Templates & 2.077\% & 1.571\% & 1.072\% & 0.298\% & 0.056\% \\
\hline
N3 Templates & 4.070\% & 3.679\% & 1.531\% & 0.894\% & 0.222\% \\
\hline
N4 Templates & 16.635\% & 15.449\% & 13.323\% & 12.071\% & 1.832\% \\
\hline
N5 Templates & 76.316\% & 78.699\% & 83.997\% & 86.662\% & 97.834\% \\
\hline
\end{tabular}
\caption{Distribution of grammatical templates of level N1(hard)-N5(easy)}
\label{tab:percentage}

\vspace{0.15in}

\begin{tabular}{|c|r|r|r|r|r|}
\hline
& N1 Texts & N2 Texts& N3 Texts& N4 Texts & N5 Texts \\
\hline
N1 Templates &3.536&2.342&0.295&0.230&0.146\\
\hline
N2 Templates &8.141&6.110&4.130&0.922&0.146 \\
\hline
N3 Templates &15.954&14.308&5.900&2.765& 0.582\\
\hline
N4 Templates &65.214&60.081&51.327 &37.327& 4.803\\
\hline
N5 Templates &299.178&306.059&323.599&267.972& 256.477\\
\hline
\end{tabular}
\caption{Number of templates of level N1(hard)-N5(easy) per 100 sentences}
\label{tab:sentence100}

\end{table*}

\end{CJK}

\subsection{Difficulty Evaluation Standard}
To evaluate the difficulty of texts and grammatical templates, we follow the standard of the Japanese-Language Proficiency Test (JLPT). The JLPT is the most widely used test for measuring proficiency of non-native speakers, with approximately 610,000 examinees in 62 countries and areas worldwide in 2011\footnote{http://www.jlpt.jp/e/about/message.html}. It has five different levels, ranging from from N5 (beginner) to N1 (advanced). A summary of the levels can be found at JLPT website \footnote{http://www.jlpt.jp/e/about/levelsummary.html}.

\subsection{Grammatical Template Library}
Due to their significance in Japanese education, grammatical templates are well-studied by Japanese teachers and researches. Grammatical templates are summarized and collected for both Japanese learners (common templates) and native speakers (templates used in very formal Japanese or old Japanese). We referenced 3 books about grammatical templates for Japanese learners~\cite{Sasaki2010JLPT,Xu2015JLPT,Liu2012JLPT}, all of which divide their templates into N1-N5 levels, for generating our template library at each corresponding level.

Although not common, books may have different opinions on the difficulty of the same template. For example, an N1 template in book A may be recognized as an N2 template in book B. In order to conduct our experiments on a reliable template library, we only pick the templates recognized as the same level by at least two of the three books. For example, if both book A and C recognized template $t$ as an N3 template, we can incorporate template $t$ into our N3 template library. Ultimately, we collected 147 N1 templates, 122 N2 templates, 74 N3 templates, 95 N4 templates and 128 N5 templates in our library. All selected grammatical templates are stored in the format of regular expressions for easy matching in parse trees. 

\subsection{Grammatical Template Extraction}
The framework of grammatical template extraction is shown in Algorithm \ref{alg:gpa}. The program requires the dependency-based parse tree of a sentence as input, runs from bottom to top and returns a set of all identified grammatical templates $\mathbf{T}(node_0)$. Line \ref{alg:t2} extracts the templates in the children of $node_0$ (and ignores the descendants of the children), by matching the phrase associated with the child nodes $[node_1,node_2,\cdots]$ to all templates stored in terms of regular expressions in our library. The matching is based on both the structure of the phrases and the properties of the words. Line \ref{alg:t12} shows $\mathbf{T}(node_0)$ covers all templates identified in subtrees rooted at $node_0$'s children and the templates extracted in the phrase associated with the child nodes $[node_1,node_2,\cdots]$.

\begin{algorithm}
\renewcommand\baselinestretch{1.3}\selectfont
\begin{algorithmic}[1]
\REQUIRE A dependency-based parse tree of the sentence
\ENSURE $\mathbf{T}(node_0) = $ set of identified grammatical templates in (sub)parse tree rooted at $note_0$.
\IF{$node_0$ is leaf node}
\STATE return $\mathbf{T}(node_0) = \{\}$
\ENDIF
\STATE $node_1,node_2,\cdots \gets$ children of $node_0$
\STATE Calculate: $\mathbf{T}(node_1),\mathbf{T}(node_1),\cdots$ // templates identified in subtrees rooted at $node_0$'s children
\STATE $\mathbf{T_1}(node_0)\gets \mathbf{T}(node_1)\cup \mathbf{T}(node_2) \cup \cdots$\label{alg:t1}
\STATE $\mathbf{T_2}(node_0)\gets$ identified templates in phrase $[node_1,node_2,\cdots]$\label{alg:t2}
\STATE return $\mathbf{T}(node_0) = \mathbf{T_1}(node_0)\cup \mathbf{T_2}(node_0)$\label{alg:t12}
\end{algorithmic}
\caption{Grammatical Progression Extraction}
\label{alg:gpa}
\end{algorithm}


We use Cabocha~\cite{kudo2002japanese} for parsing Japanese sentences. This tool generates the hierarchical structure of the sentence as well as some properties (e.g. base form, pronunciation, part of speech, etc.) of each word. We execute Algorithm \ref{alg:gpa} on the parse tree to extract all identified templates of a Japanese sentence.

\section{Statistics of Grammatical Templates}

\subsection{Corpus}
We build our corpus from two sources: past JLPT exams and textbooks. The reading texts from JLPT exams are ideal for difficulty evaluation experiments since all of them are tagged authoritatively with difficulty levels, and JLPT problem sets before 2010 are publicly released\footnote{For example, the second exam in 2009 is published in~\cite{JLPT2009}.}. We also collected reading texts from two popular series of Japanese textbooks: \emph{Standard Japanese}~\cite{StandardJapanese} and \emph{Genki}~\cite{Genki}. \emph{Standard Japanese I} and \emph{Genki I} are designed for the N5 level (the first semester) and \emph{Standard Japanese II} and \emph{Genki II} are designed for the N4 level (the second semester). Ultimately, our corpus consists of 220 texts (150 from past JLPT exams and 70 from textbooks), totaling 167,292 words after segmentation.

\subsection{Results}
For texts with different difficulties, we calculate the distribution of N1-N5 grammatical templates, which are shown in Table \ref{tab:percentage}. We can see that N1 texts have higher portion of N1 and N2 templates than N2 texts, implying that the difficulty boosts from N2 to N1 are derived from increasing usage of advanced grammar. It is also clear that even in the texts of advanced levels, the majority of the sentences are organized by elementary grammatical templates, and the advanced ones are only used occasionally for formality or preciseness.

We also calculate the per-100-sentence number of templates at each level, which are shown in Table \ref{tab:sentence100}. When comparing any two adjacent levels (e.g. N2 and N3), the templates at those levels or above seem to be the most significant. For instance, N1/N2 texts differ in numbers of N1 and N2 templates while they have similar numbers of N3-N5 templates, and the numbers of N1, N2 and N3 templates differentiate the N2/N3 texts while the numbers of N4 and N5 templates seem relatively similar. This phenomenon inspires us to build a simple and effective approach to differentiate the texts of two adjacent levels.

\section{Difficulty Level Prediction}

\subsection{Multilevel Linear Classification}
\label{sec:mlc}
We differentiate two adjacent levels by looking at the knowledge `on the boundary' and `outside the boundary'. Concretely, when judging whether a text is harder than level $N_i$, we consider a grammatical template as:

\begin{itemize}
\item \emph{within the boundary}, if the template is easier than $N_i$ ($N_{i+1}$ to $N_5$);
\item \emph{on the boundary}, if the template is exactly at $N_i$ level;
\item \emph{outside the boundary}, if the template is harder than $N_i$ ($N_1$ to $N_{i-1}$).
\end{itemize}

\begin{figure}
\centering
\includegraphics[width=0.8\textwidth]{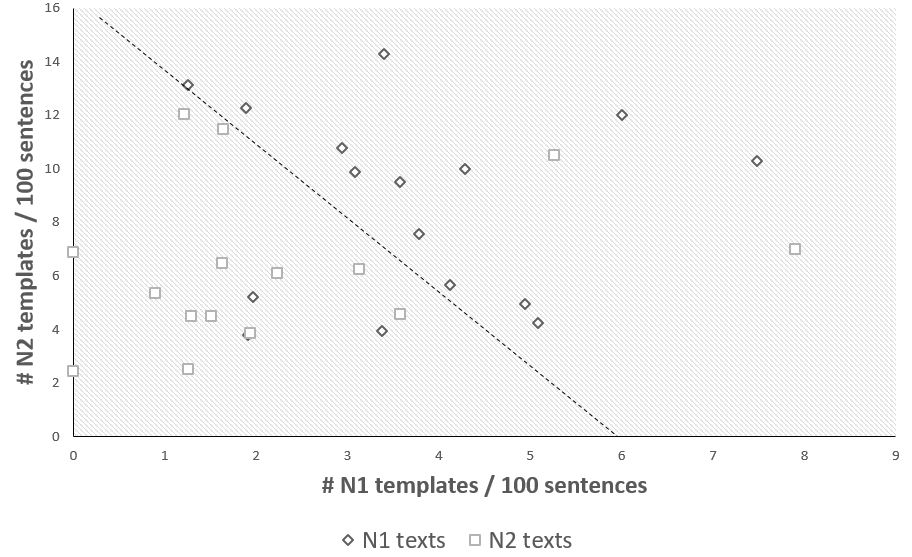}
\caption{Grammatical difficulty in the N1/N2 texts}
\label{fig:N1N2}

\vspace{0.2in}

\centering
\includegraphics[width=0.8\textwidth]{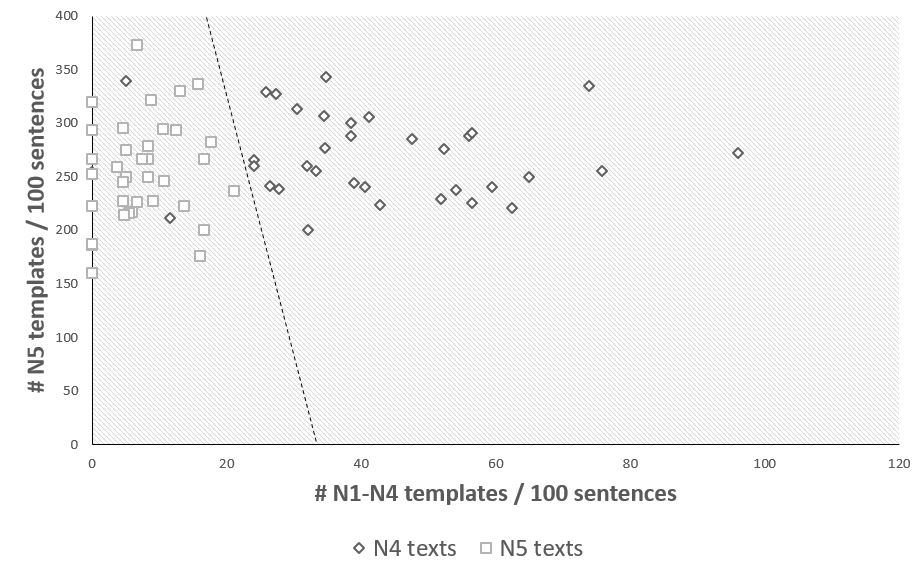}
\caption{Grammatical difficulty in the N4/N5 texts}
\label{fig:N4N5}

\vspace{0.4in}

\centering
\scalebox{1.1}{
\tikzstyle{decision} = [diamond, draw, fill=lightgray!20, 
    text width=3em, text badly centered, node distance=2cm, inner sep=0pt]
\tikzstyle{block} = [rectangle, draw, fill=lightgray, 
    text width=2.5em, text centered, rounded corners, minimum height=1em, node distance=1.5cm]
\tikzstyle{finalblock} = [rectangle, draw, fill=lightgray, 
    text width=2.5em, text centered, rounded corners, minimum height=1em, node distance=2cm]
\tikzstyle{line} = [draw, -latex']
\tikzstyle{cloud} = [draw, ellipse, node distance=1cm,
    minimum height=1em]
    
\begin{tikzpicture}[node distance = 1.5cm]
    \node [cloud] (text) {Text};
    \node [decision, right of=text] (n5c) {$>$N5?};
    \node [block, below of=n5c] (n5) {N5};
    \node [decision, right of=n5c] (n4c) {$>$N4?};
    \node [block, below of=n4c] (n4) {N4};
    \node [decision, right of=n4c] (n3c) {$>$N3?};
    \node [block, below of=n3c] (n3) {N3};
    \node [decision, right of=n3c] (n2c) {$>$N2?};
    \node [block, below of=n2c] (n2) {N2};
    \node [finalblock, right of=n2c] (n1) {N1};
    \path [line] (text) -- (n5c);
    \path [line] (n5c) -- node [midway,right]{no} (n5);
    \path [line] (n5c) -- node [midway,above]{yes} (n4c);
    \path [line] (n4c) -- node [midway,right]{no} (n4);
    \path [line] (n4c) -- node [midway,above]{yes} (n3c);
    \path [line] (n3c) -- node [midway,right] {no} (n3);
    \path [line] (n3c) -- node [midway,above]{yes} (n2c);
    \path [line] (n2c) -- node [midway,right]{no} (n2);
    \path [line] (n2c) -- node [midway,above]{yes} (n1);

\end{tikzpicture}
}
\caption{Multilevel Linear Classification (MLC). `$>$N5?' represents the linear classifier judging whether a text is harder than N5. The classifiers are similar for the other levels.}
\label{fig:mlc}
\end{figure}

We found that texts of adjacent levels are nearly linear-separable with two features: 
templates `on the boundary' and templates `outside the boundary'. For example, Figure \ref{fig:N1N2} shows how N1 and N2 texts are linearly separated based on the numbers of N1 and N2 templates: we can easily obtain a two-dimensional linear classifier separating N1 and N2 texts with $83.4\%$ accuracy.  This phenomenon is even more obvious at lower levels. Figure \ref{fig:N4N5} shows N4 and N5 texts are almost perfectly linearly separated with two features: `number of N5 templates per 100 sentences' (on the boundary) and `number of N1-N4 templates per 100 sentences' (outside the boundary).

Taking advantage of this phenomenon, we build 4 linear classifiers for 4 pairs of adjacent levels. For example, the N4 classifier judges whether a text is harder than N4 (N1-N3). Our \emph{Multilevel Linear Classification (MLC)}  algorithm combines all 4 linear classifiers: A text is judged by the N5 classifier first. If it is no harder than N5, it will be labeled as an N5 text; otherwise, it will be passed to the N4 classifier in order to decide if it is harder than N4. The process continues similarly, until if it is judged to be harder than N2, it will be labeled as an N1 text. Figure \ref{fig:mlc} shows how the algorithm works.

\subsection{Features}

We conduct our experiments on the following 4 feature sets:

First, our \emph{grammatical template feature set} has only 5 features: 

\begin{itemize}

\item Average number of N1-N5 grammatical templates per sentence
\end{itemize}

We compare our work with recent readability evaluation studies~\cite{kim2014device,pitler2008revisiting}. In our experiments, the \emph{baseline readability feature set} consists of the following 12 features:
\begin{itemize}
\itemsep-0.03in
\item Number of words in a text
\item Number of sentences in a text
\item Average number of words per sentence
\item Average parse tree depths per sentence
\item Average number of noun phrases per sentence
\item Average number of verb phrases per sentence
\item Average number of pronouns per sentence
\item Average number of clauses per sentence
\item Average cosine similarity between adjacent sentences
\item Average word overlap between adjacent sentences
\item Average word overlap over noun and pronoun only
\item Article likelihood estimated by language model
\end{itemize}

Moreover, we combine these 12 traditional readability features with our 5 grammatical template features, forming a \emph{`hybrid' feature set}, since we would like to see if grammatical template features are really able to improve text difficulty evaluation.

Since the text difficulty level prediction can be regarded as a special text classification problem, we also extract \emph{TF-IDF features}~\cite{sparck1972statistical} ~\cite{nelson2012measures} as an extra baseline, in order to see how general text classification techniques work on text difficulty evaluation.

\subsection{Result}
We test k-Nearest Neighbor and Support Vector Machines for each feature set. The implementations of these two popular classification algorithms are provided by the WEKA toolkit~\cite{hall2009weka} and LibSVM~\cite{chang2011libsvm}. The SVMs use RBF kernels~\cite{chang2010training}. We also test our Multilevel Linear Classification (MLC) algorithm on the grammatical template feature set. We use 5-fold cross validation to avoid overfitting. Table \ref{tab:accuracy} shows the results.

\begin{table}
\centering
    \begin{tabular}{|c|c|c|}
    \hline
    \multirow{2}[0]{*}{Feature Set (number of features)} & \multirow{2}[0]{*}{Algorithm} & \multirow{2}[0]{*}{Accuracy}\\
    &&\\
    \hline
    \multirow{2}[0]{*}{TF-IDF Features (5100)} & kNN   & 69.1\% \\
    & SVM   & 80.5\% \\
    \hline
    \multirow{2}[0]{*}{Baseline Readability Features (12)} & kNN & 72.3\%\\
    & SVM   & 80.9\% \\
    \hline
    \multirow{3}[0]{*}{Grammatical Template Features (\textbf{5})} & kNN & 78.0\%\\
    & SVM   & 81.1\% \\
    & \textbf{MLC}   & \textbf{87.7\%} \\
    \hline
    \multirow{2}[0]{*}{Hybrid Features (17)} & kNN & 85.7\%\\
    & SVM   & \textbf{88.5\%} \\
    \hline
    \end{tabular}%

\caption{Accuracies of classifying N1-N5 texts}
\label{tab:accuracy}
\end{table}

Comparing the results of kNN and SVM across the four different feature sets in Table~\ref{tab:accuracy}, it is clear that TF-IDF features have the largest feature set yet lowest accuracy, indicating the general word-based text classification techniques do not work well on text difficulty level prediction. Compared with baseline readability features, our grammatical template features have smaller number of features but higher accuracy (slightly higher with SVM but significantly higher with kNN). Moreover, the hybrid features, which combine baseline readability features with grammatical template features, decisively outperform baseline readability features, confirming our expectation that adding grammatical template features to existing readability techniques improves text difficulty evaluation for language learners.

Additionally, our Multilevel Linear Classification algorithm achieves excellent accuracy with only 5 grammatical template features. An accuracy of 87.7\% , although slightly lower than hybrid features + SVM (more features, more complexity), still significantly outperforms baseline readability techniques. In conclusion, the Multilevel Linear Classification algorithm has high accuracy, a small number of features, and a simple, human-understandable structure.

\section{Conclusions and Future Work}

We proposed a new approach for evaluating text difficulty that focuses on grammar and utilizes expert-identified grammatical templates, the grammar knowledge that students actually learn in language lessons. This approach significantly improved the accuracy of text difficulty evaluation for Japanese language learning. We also introduced a simple, human-understandable, and effective text difficulty evaluation approach using only five grammatical template features. 


In future work, we are interested in extending our work to other languages like English, and adapting grammatical templates for various languages. To achieve this, we need to itemize the grammar knowledge that students learn from language lessons. We can also develop a machine learning system that can automatically discover discriminative grammatical templates from texts. Moreover, we would like to study if the topic of a text has considerable impact on text difficulty for language learners, just like vocabulary and grammar.

We also hope to use our approach to recommend reading texts to individual learners at appropriate difficulty levels. For instance, Japanese news articles could be good learning materials for advanced Japanese language learners. We want to build an online tool to collect reading texts from current news reports in specific target languages, and select appropriate ones for language learners, especially intermediate and advanced learners.

Finally, we plan to leverage some novel ideas from Human-Computer Interaction and educational technology~\cite{andersen2013trace} to build an \emph{adaptive} Computer-Assisted Language Learning (CALL) system. Using our new approach introduced in this paper, we can decompose the difficulty of a text into several basic skills (grammatical templates), and model the internal hierarchical structure of a sequence of texts with a partial ordering graph. Using this structure, we can comprehensively assess a student's knowledge and tailor optimal learning progressions for individual students.




\section*{Acknowledgements}


Special thanks to Xiang Long for his help during the writing of this paper.


\bibliography{bib}
\bibliographystyle{acl}

\end{document}